\documentclass[letterpaper]{article}
\usepackage{aaai}
\usepackage{times}
\usepackage{helvet}
\usepackage{courier}
\usepackage{natbib}
\usepackage{algorithmic}
\usepackage{graphicx}
\title{Code Representation Learning Using Pr{\"u}fer Sequences}
\author{Tenzin Jinpa and Yong Gao \\
Department of Computer Science \\
University of British Columbia, Okanagan \\
Kelowna, BC, Canada \\
tenzin.jinpa@ubc.ca , yong.gao@ubc.ca 
}
\begin{document}
\maketitle
\begin{abstract}
\begin{quote}
An effective and efficient encoding of the source code of a computer program is critical to the success of sequence-to-sequence deep neural network models for tasks in computer program comprehension, such as automated code summarization and documentation. A significant challenge is to find a sequential representation that captures the structural/syntactic information in a computer program and facilitates the training of the learning models. 

In this paper, we propose to use the Pr{\"u}fer sequence of the Abstract Syntax Tree (AST) of a computer program to design a sequential representation scheme that preserves the structural information in an AST. Our representation makes it possible to develop deep-learning models in which signals carried by lexical tokens in the training examples can be exploited automatically and selectively based on their syntactic role and importance. Unlike other recently-proposed approaches, our representation is concise and lossless in terms of the structural information of the AST. Empirical studies on real-world benchmark datasets, using a sequence-to-sequence learning model we designed for code summarization, show that our Pr{\"u}fer-sequence-based representation is indeed highly effective and efficient, outperforming significantly all the recently-proposed deep-learning models we used as the baseline models.      
         
\end{quote}
\end{abstract}

\begin{figure*}
  \includegraphics[width=\textwidth,height=10cm]{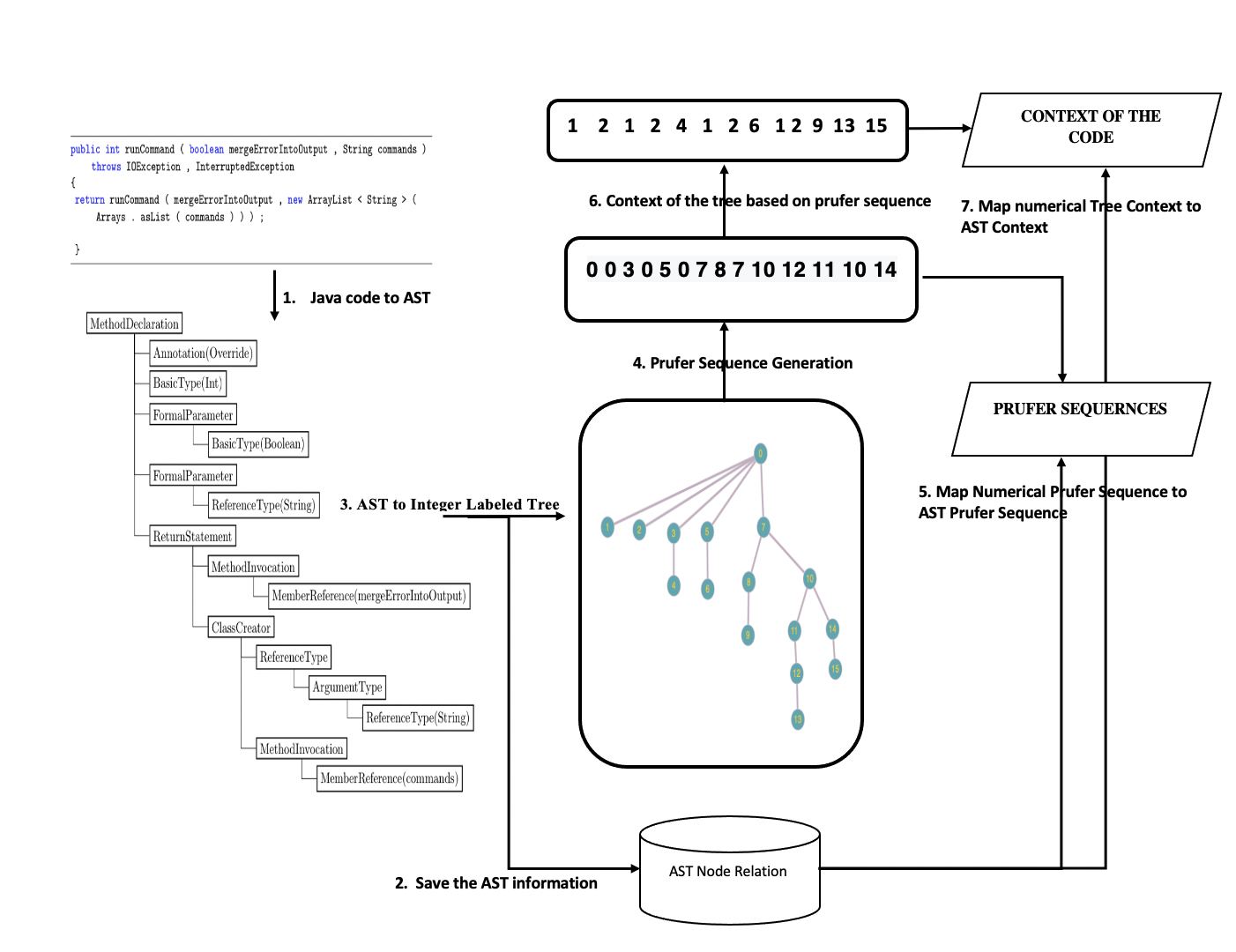}
 \caption{Pr{\"u}fer Sequences and Context of an AST}
\end{figure*}


\section{Introduction}
The use of deep learning techniques and related neural machine translation (NMT) models in code understanding has drawn much recent attention. It has been shown that methods using  NMT models can achieve a much better performance than traditional Informational Retrieval (IR) techniques in tasks such as automated code summarization (\citealt{a1,a3}) and program property prediction (\citealt{a4}), thus improving productivity (\citealt{b2}) and reducing software development costs (\citealt{b3}). Unlike natural languages, which are unstructured and noisy, computer programs are highly structured. It is thus critical to encode as much as possible the structural information in a sequence-to-sequence learning model and to take advantage of the encoded information in the training.

Recently, several methods have been proposed to incorporate the structural information in the Abstract Syntax Tree (AST) of a computer program in a sequence-to-sequence learning model. In the Structure-Based Traversal (SBT) method proposed by \cite{a5}, an AST is represented by a sequence of syntactic tokens and is generated by a depth-first traversal of the AST with parentheses pairs to retain the sub-tree information. The model Code2Seq (\citealt{a4}) uses the concatenation of the token sequences along the paths between pairs of terminal nodes in an AST to represent a computer program.  While these methods and models have been shown to be effective in comparison to the most straightforward representation by ``flat" sequence of tokens arranged in the same order as they appear in the source code, the choices of the traversal method and the ordering of the tokens appear to be still arbitrary. For example,  for a standard tree traversal method such as the depth-first search, there are at least three ways to order the tokens in the sequence: in-order, pre-order, and post-order. We also note that the length of these sequence representations (and thus the input dimension of the learning model) is significantly increased.  Due to the use of parentheses for sub-tree information, the length of the representation proposed by \cite{a5} is, in the worst case, three times as long as that of the source code. The length of the representation used in the model Code2Seq (\citealt{a4}) is, in the worst case, cubic in the length of the source code.

In this paper, we propose to use the Pr{\"u}fer sequence of the Abstract Syntax Tree (AST) of a computer program to design a sequential representation scheme that preserves the structural information in an AST. Our representation makes it possible to develop deep-learning models in which signals carried by lexical tokens in the training examples can be exploited automatically and selectively based on their syntactic role and importance. Unlike other recently-proposed approaches, our representation is concise and lossless due to the fact that an AST can be uniquely reconstructed from its Pr{\"u}fer sequence. Empirical studies on real-world benchmark datasets,  using a sequence-to-sequence learning model we designed for code summarization, showed that our Prufer-sequence-based representation is indeed highly effective and efficient, and our model outperforms significantly all the recently-proposed deep-learning models used as the baselines in the experiments.  

\section{Previous Work}

The application of deep learning models in representation learning for computer programs has attracted much recent attention and is widely used for many tasks in computer program comprehension such as automated code summarization (\citealt{a7,a8,a9}), code generation (\citealt{a10}), and code retrieval (\citealt{a11}).


Several approaches have been investigated to making use of syntactic and structural information, explicitly or implicitly, in representation learning from the source code of computer programs. \cite{a12} used the relations in the abstract syntax tree as a feature for training a learning model. \cite{a13} and \cite{a14} used the paths in an AST to identify the context node. \cite{a5} proposed a Structure-Based Traversal (SBT) method to represent ASTs as a linear sequence containing syntactic information in a sequence-to-sequence learning model for code summarization. \cite{a15} further extended their SBT-based model (\citealt{a5}) by adding to their model another encoder that learns from lexical information in the source code. Another representation (\citealt{a16}) uses the concatenation of the token sequences along the paths between pairs of terminal nodes in an AST. In addition to these efforts of directly using the structural information in the ASTs in a sequence-to-sequence model, \cite{a9} explored the possibility of exploiting 
structural information implicitly with  
a transformer model enhanced by pariwise semantic relationships of tokens
in the model's attention mechanism. 
A significant downside of a transformer-based approach is the increase of the model complexity (quadratic in the code length), and thus the sample complexity and the computational complexity.

\section{Abstract Syntax Trees and Pr{\"u}fer Sequences}

Pr{\"u}fer sequences have been used in the past as a sequential representation of tree structures in stochastic search methods (\citealt{pru-r1,pru-r2}), problems in fuzzy systems (\citealt{pru-r3,pru-r5}), and hierarchical or graph data management (\citealt{pru-r7,pru-r8}). To the best of our knowledge, our work is the first effort in using  Pr{\"u}fer sequences and exploiting their unique properties in (deep) representation learning for structured data. 

In this section, we discuss the details of our Pr{\"u}fer-sequence-based code representation, including its main idea and its construction. We also discuss the advantages of our representation over other sequential representations in terms of effectiveness, efficiency, and flexibility.

\subsection{ASTs and Sequence to Sequence Models }

An abstract syntax tree (AST) is a tree structure that models the abstract syntactic structure of the source code of a computer program. In an AST, the leaves (or terminal nodes) are labeled by tokens that contain user-defined values or reference types such as variable names, whereas internal nodes (or non-terminal nodes) are labeled by tokens that summarize the purpose of code blocks such as conditions, loops, and other flow-control statements. 
We note that a token labeling an internal node does not have to be from the source code or specification of the particular programming language. 
By slightly abusing the notion, we call a token labeling a leave node a \textbf{``lexical token"} as it contains program-specific information in the source code and call a token labeling a non-terminal node a \textbf{``syntactic token"} as it contains generic information about the structure and the purpose of a code block.  Shown in Fig.1 is a function in Java and its AST where the token set 
$\{${\it Override, Int, String, mergeErrorIntoOutput, Boolean, Commands}$\}$ are used to label the terminal nodes and the token set $\{${\it BasicType, FormalParameter, MethodInvocation, ReturnStatement, ClassCreator, ReferenceType}$\}$ are used in label the internal nodes.

In a sequence-to-sequence model for code representation learning, tokens (or their word embedding) from the source code have to be represented as a linear sequence and are used as the input of the model. As has been shown by \cite{a5} and \cite{a16}, a linear sequence representation that makes use of structural information encoded in the AST has a great advantage over a simple flat sequence representation where tokens appear in the same order as they appear in the source code (\citealt{a6}).  

\subsection{Pr{\"u}fer Sequence of an AST }

The Pr{\"u}fer sequence of a node-labeled tree is a sequence of node labels from which the tree can be uniquely reconstructed. The famous proof of Caley's formula for the number of labeled trees by Heinz Pr{\"u}fer (\citealt{a18,z1}) is based on the one-to-one correspondence between the set of labeled trees and the set of such sequences. Given a tree $T$ with $n$ nodes labeled by the integers $\{1, \cdots, n\}$, its Pr{\"u}fer sequence is a sequence of $(n - 2)$ node labels (i.e., integers) and can be formed by successively removing the leave with the smallest label and including the label of its parent as the next node label in the Pr{\"u}fer sequence. The process stopped when only two nodes were left in the tree. 

Since ASTs are labeled by syntactic and lexical tokens,  we use a fixed mapping
to map each token in the given token set to a unique integer and use it as the integer label of the AST-node that is labeled by the token.  The Pr{\"u}fer sequence constructed from this integer-labeled AST is then mapped back to a sequence of syntactic tokens, which we call the \textbf{``syntactic Prufer sequence"} and is used as part of the input sequence to our learning model. Fig. 1 illustrates the process of Pr{\"u}fer sequence generation from the AST of a Java method. The syntactic Pr{\"u}fer sequence for the Java method in Fig. 1 is   

\vspace{3pt}

$\{${\it MethodDeclaration, MethodDeclaration, FormalParameter, MethodDeclaration, FormalParameter, MethodDeclaration, ReturnStatement, MethodInvocation, MethodInvocation, ClassCreator, TypeArgument, ReferenceType, ClassCreator, MethodInvocation}$\}$ 
 
Note that in the above sequence, a syntactic token may appear multiple times and at different positions. Also, note that the terminal nodes
never appear in the sequence.  The significance and relevance of those properties of the syntactic Pr{\"u}fer sequence will be discussed below and further explored in the next two sections on the design of our learning model and our empirical studies.

\begin{figure*}
  \includegraphics[width=\textwidth,height=7cm]{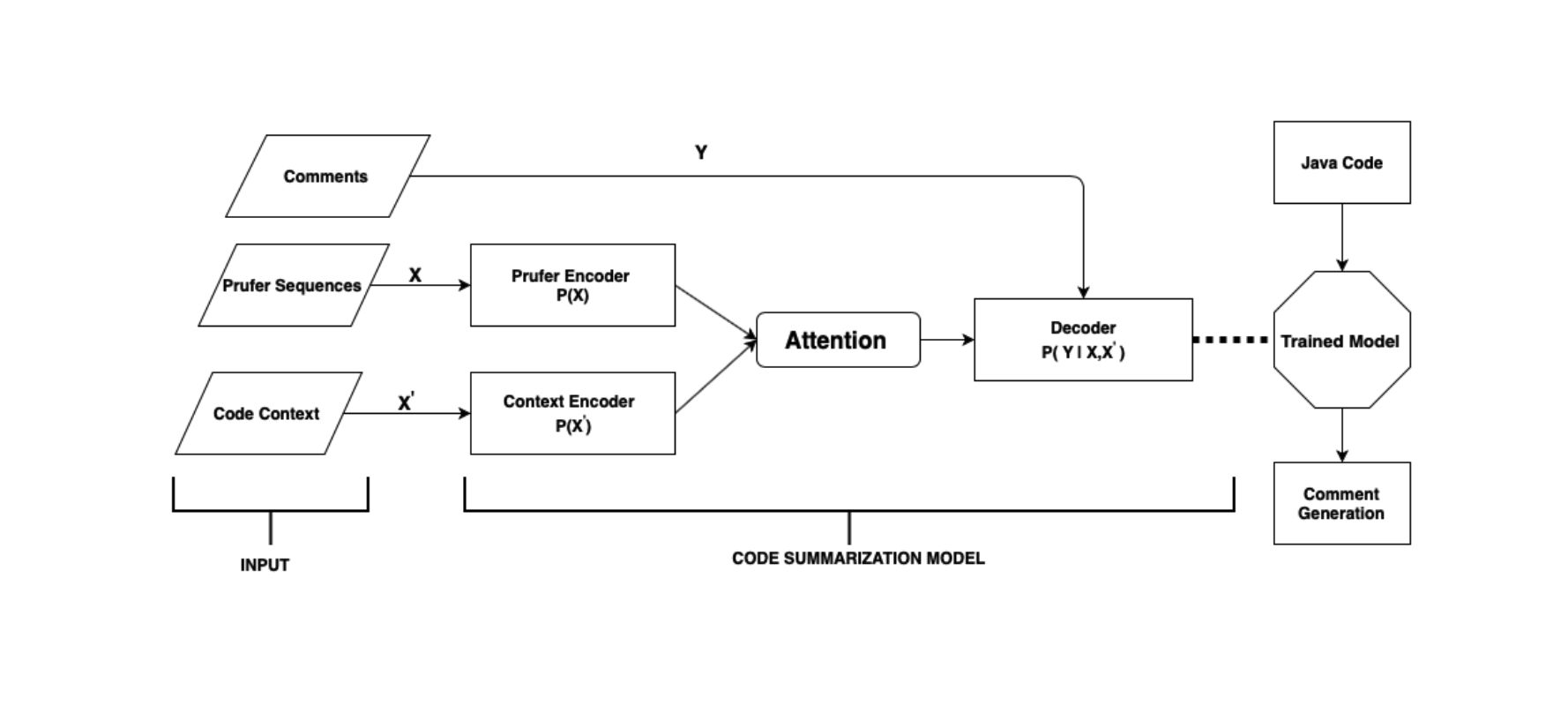}
 \caption{Pr{\"u}fer Based Learning Model for Code Summarization}
\end{figure*}


\subsection{Advantages of Learning with a Pr{\"u}fer-Sequence Representation}

The syntactic Pr{\" u}fer sequence can be regarded as a ``transformed" and ``quantified" version of an AST and the corresponding source code where     
\begin{enumerate}
   \item the frequency with which a syntactic token appears is decided by the degree of the corresponding AST node and quantifies the ``importance" of the token (measured the size of the code block it controls);
    
    \item the positions of the appearances of syntactic tokens in the  Pr{\"u}fer sequence are decided by the position of the corresponding node in the tree; and 
    
    \item a lexical token labeling a leave node of an AST never appears in the  Pr{\"u}fer sequence, but its ``significance" can be measured by the syntactic importance of the parent of the leave node.
    
\end{enumerate}
This is in sharp contrast with all other recently proposed sequential representations of a source code and its AST, where all tokens are treated equally, and their positions only partially capture their roles in the AST.              

Supported by observations from our empirical studies (discussed in the second last section), we believe that these are the properties that make it possible (or much easier) for our learning models to exploit information in the training examples that are hard, if not impossible, for other recently-proposed learning models to detect.          
                    
As we observed in our experiments,  other properties that distinguish our Prufer-sequence representation from exiting representations and play important roles in the performance of our learning model are as follows.  
\begin{enumerate}

\item A Uniqueness and Lossless Representation \newline 
To our best knowledge, all existing sequential representations of source code and its AST  (\citealt{a4,a5}) are lossy encoding in the sense that the original AST cannot be uniquely reconstructed.  Our Pr{\"u}fer-sequence representation is a lossless encoding because, given a fixed syntactic-token-to-integer mapping, there is a one-to-one correspondence between the set of ASTs and their syntactic Pr{\"u}fer sequences. This property may help improve the ability of a learning method to distinguish or detect subtle differences in training examples.

\item More Concise Input (or Lower Input Dimension)\newline For an AST with $n$ nodes,  the length of our representation is  
$n - 2$. In comparison, the length of the representation proposed by \cite{a5} is $3n$ in the worst case,  while the length of the representation proposed by \cite{a16} is in $\Omega$ $(n^3)$ in the worst case.  In our experiments, we observed that the Pr{\"u}fer sequence representation has an average length of 100.81 tokens while the representation based on SBT (\citealt{a5}) has the 193.71 tokens to represent the same AST corpus (Table 5), resulting in faster training of our model. 

\end{enumerate}

\section{Pr{\"u}fer-Sequence-Based Learning Model For Code Summarization }

To study the effectiveness of the Pr{\"u}fer-sequence-based representation, we developed a deep-learning model for code summarization. The model maps a Java method to a summary of the method's purpose in English. The training data are pairs of source code of the Java method and developers' comments.  The high-level structure of our model is depicted in Fig.2. It is a sequence-to-sequence (seq2seq) model (\citealt{a19}) in the encoder-decoder paradigm, where two separate encoders are used to learn from the structural information of an AST and from a structure-aware representation of the lexical tokens from the source code. An attention module, similar to the one used by \cite{a15}, is used to combine the output of the two encoders into a context vector which is then used as the input to a standard decoder described in (\citealt{a19}) to output a code summary/comment in English.     

We describe the details of the two encoders and their rationale below.

\subsection{Pr{\"u}fer Sequence Encoder}

The Pr{\"u}fer Sequence Encoder is designed to learn from the structural 
information of the ASTs that are losslessly encoded in their syntactic Pr{\"u}fer sequences. Gated Recurrent Units (GRUs), as discussed by (\citealt{a21}), are used  
to map the syntactic Pr{\"u}fer sequence 
$(X = x_1, ....,x_n)$ of a computer program to a sequence 
of hidden states as follows:     
      \begin{center}
      $ s_t = GRU(x_t,s_{t - 1})$
       \end{center} 

In our implementation, lexical tokens labeling the terminal nodes are appended to the syntactic Pr{\"u}fer sequence as part of the input to the encoder, resulting in an input length of at most $2n - 3$. We observed in our initial experiments that the model Hybrid-DeepCom (\citealt{a15}) with its SBT-based encoder replaced by our Pr{\"u}fer Sequence Encoder already outperformed notably the original Hybrid-DeepCom Model and other baseline models. It turned out that the bottleneck to further improvement of such models is the design of the second encoder, the Source Encoder (\citealt{a15}), that learns from source code tokens directly. This observation in our early investigation motivated us to design our own second encoder, which we call the Context Encoder, that exploits lexical information in a structure-aware way. We discuss our definition of the context of an AST and the design of the Context Encoder in the next subsection.

\subsection{Context Encoder} 
The context encoder, also consisting of GRUs, is designed to learn 
from the collection of lexical tokens (i.e., user-defined and program-specific values in the source code) organized in a way that reflects the structural information of the AST. 

For each node in an AST, we define its {\bf context} to be the set of lexical tokens that label the node's leaf child/children. A node with no leaf child has an empty context. The context of an AST is defined to be the union of the contexts of the AST nodes ordered in the same way as they appear in the syntactic Pr{\"u}fer sequence. The context of an AST 
can be calculated from the AST's Pr{\" u}fer sequence (See Fig. 1).
The context encoder maps the context defined in the above to a sequence of hidden states.

The context of an AST defined in this way is a \textbf{structure-aware} sequence of lexical tokens because (1)  the frequency of a lexical token is decided by the degree of the parent of the leaf node that the token labels; and (2) the order in which these  tokens appear in the context is the same as the order of the parent nodes in the Pr{\"u}fer sequence.   
As observed in our experiments, the use of the context encoder helps boost the performance of our learning model significantly.  This is because, we believe, that the context we have designed helps amplify the learning-relevant lexical signals in the source code in a way that other models (such as those in \cite{a15}) cannot detect that simply use the collection of entire tokens as they appear in the source code.      
    
\section{Experimental Studies}

In this section, we first introduce the experimental setup, including the datasets, the baseline models, and the evaluation metrics. 
We then discuss observations from and analyses of our experimental results on the power, effectiveness, and efficiency of the proposed Pr{\" u}fer sequence representation and the role it plays for our learning model to significantly outperform recently-proposed learning models.

\subsection{Dataset and Experiment Setup}

We perform our experiment on two Java datasets. Dataset-1(68469 pairs of java code and comments) is a popular dataset used in many research and is collected by \cite{a23} from popular repositories in Github. Dataset-2 ( 163316 pairs of java code and comments) is part of the CodeXGlue dataset developed by Microsoft (\citealt{codexglue}). It is known for its high quality and complexity and is believed to be one of the most challenging datasets for deep learning approaches to program understanding and generation.  We split the data into 8:1:1 for training, testing, and validation. Some statistics of the datasets are shown in Tables 1 and 2. 

We tokenized the java source code and comments by the programs Javalang and NLTK\footnote{https://www.nltk.org/} respectively. The size of the vocabulary for comments, code, code's context, and  Pr{\"u}fer sequence is set to 30000 (\citealt{a5,a19}). The maximum length of the Pr{\"u}fer sequence was kept at 200, and the size of the code and code's context was kept at 500 (\citealt{a5,a19}). Special tokens $<$START$>$ representing the start of the sequence and $<$EOS$>$  representing the end of the sequence are added to the decoder sequence during the training. The maximum comment length is set to 30  and out-of-vocabulary represented by especial token $<$UNK$>$. 

Our model uses one layered GRU with 256 dimensions of hidden state and 256-dimensional word embedding.  The maximum iterations are 60 epochs. The learning rate is set to 0.5, and we clip the gradients norm by 5. The learning rate is decayed using the rate 0.99. The model uses the TensorFlow version 1.15, and we train our model on a single GPU of Tesla P100-PCIE-16GB with 25 GB RAM and 110GB disk. 

\vspace{-1em}
\begingroup
\setlength{\tabcolsep}{2pt} 
\renewcommand{\arraystretch}{1} 
\begin{table}[h!]
\caption{Statistics of the Java Methods}
  \begin{center}
    \begin{tabular}{l|c|c|r|r} 
       \textbf{Dataset Type}&\textbf{Avg.} & \textbf{Mode} & \textbf{Median}   & \textbf{$<$200 Tokens} \\
      \hline
        Dataset-1 & 97.05 & 16 & 64  & 89.48\% \\
        Dataset-2 & 98.35 & 42 & 69  & 89.29\% \\
       
 \hline
    \end{tabular}
    \vspace{-2em}
  \end{center}
 \end{table}

\endgroup

\begingroup
\setlength{\tabcolsep}{2pt} 
\renewcommand{\arraystretch}{1} 
\begin{table}[h!]
\caption{Statistics of the Java comments}
  \begin{center}
    \begin{tabular}{l|c|r|r|r} 
    
       \textbf{Dataset Type} &\textbf{Avg.} & \textbf{Mode} & \textbf{Median} & \textbf{$<$30 Tokens}  \\
      \hline
     Dataset-1 & 13 & 7 & 11 & 88.42\%\\
     Dataset-2 & 11.02 & 7 & 9  & 97.33\%\\
       
 \hline
    \end{tabular}
    \vspace{-2em}
    
  \end{center}
\end{table}
\endgroup


\begingroup
\setlength{\tabcolsep}{1pt} 
\renewcommand{\arraystretch}{1} 
\begin{table*}[h!]
 \caption{Effectiveness of Models based on Machine Translation metrics for Dataset-1}
  \begin{center}
    \begin{tabular}{l|c|r|r|r} 
      \textbf{Model} & \textbf{S-BLEU} & \textbf{C-BLEU} & \textbf{Meteor} & \textbf{ROUGE-L} \\
      \hline
      
      Lexical-Token-Only Model & 36.21 & 27.30 & 19.01 & 40.78\\
      
      Code2Seq & 20.72 & 4.56 & 10.21 & 20.63 \\
      
      TL-CodeSum & 37.20  & 28.43 & 19.64 & 41.29\\
      
      BFS-Hybrid-DeepCom & 37.98
 & 29.08 & 19.72 & 41.03\\

      Hybrid-DeepCom & 38.19 & 29.28
 & 19.87 & 41.15\\
 \hline
 \textbf{Our Model (Pr{\"u}fer Encoder + Hu's Source Encoder)} & \textbf{38.38 ($0.5\%$)}
 & \textbf{29.43 ($0.5\%$)}  & \textbf{20.13 ($1.3\%$)} & \textbf{41.82 ($1.3\%$)}\\

 \textbf{Our Model (Pr{\"u}fer Encoder + Context Encoder)} & \textbf{39.67 ($3.3\%$)}  & \textbf{31.01 ($5.7\%$)}  & \textbf{21.01 ($5.6\%$)} & \textbf{43.45 ($5.1\%$)}\\
 
 \hline
    \end{tabular}
    \vspace{-2em}
  \end{center}
\end{table*}
\endgroup

\begingroup
\setlength{\tabcolsep}{1pt} 
\renewcommand{\arraystretch}{1} 
\begin{table*}[h!]
 \caption{Effectiveness of Models based on Machine Translation metrics For Dataset-2}
  \begin{center}
    \begin{tabular}{l|c|r|r|r} 
      \textbf{Model} & \textbf{S-BLEU} & \textbf{C-BLEU} & \textbf{Meteor} & \textbf{ROUGE-L}\\
      \hline
      Lexical-Token-Only Model & 9.21 & 3.07 & 7.96 & 19.84\\
      
      Code2Seq  & 2.27 & 0.30 & 3.5 & 12.23 \\
      
      BFS-Hybrid-DeepCom & 13.41
 & 3.47 & 7.29 & 20.42\\

      Hybrid-DeepCom & 15.02 & 3.7
 & 8.27 & 18.01 \\
 \hline
\textbf{Our Model (Pr{\"u}fer Encoder + Hu's Source Encoder)} & \textbf{15.50 ($3.15\%$)}
 & \textbf{3.85 ($3.97\%$)}  & \textbf{8.9 ($6.925\%$)} &  \textbf{20.79 ($1.8\%$)}\\

 \textbf{Our Model (Pr{\"u}fer Encoder + Context Encoder)} & \textbf{16.15($7.02\%$)}  & \textbf{4.49 ($19.29\%$)}  & \textbf{9.72 ($15.05\%$)} & \textbf{ 24.73 ($19.09\%$)}\\
 
 \hline
    \end{tabular}
    
  \end{center}
\end{table*}
\endgroup


\subsection{Baseline}
In our experiments, we compared the performance of our learning model with the following baseline models to empirically analyze the power, effectiveness, and efficiency of the proposed Pr{\"u}fer-sequence-based  representation.   

\begin{enumerate}
    
\item \textbf{TL-CodeSum Model} (\citealt{a23}). This is an NMT based code summarization method that uses API knowledge and source code tokens as the input in a sequence-to-sequence model. 

\item \textbf{Hybrid-DeepCom Model} (\citealt{a15}). This sequence-to-sequence model for code summarization is one of the recent models that is designed to exploit structural information in the AST of a computer program. It uses two encoders: a source-token encoder and an SBT encoder. The SBT encoder uses a depth-first-traversal sequence representation of an AST as its input, and the code-token encoder is used to learn from the lexical information of the source code. 

\item \textbf{Code2Seq Model} (\citealt{a16}). This is a deep-learning model for general code representation learning.  Code2Seq 
uses the concatenation of the token sequences along the paths between pairs of terminal nodes in an AST as its input representation.

\item \textbf{BFS-Hybrid-DeepCom}.  This model is based on the Hybrid-DeepCom Model  (\citealt{a15}) with the SBT representation of an AST  replaced by a sequence representation constructed from a breadth-first-search (BFS) traversal of the AST. We included this model to help verify our claim that all the recently-proposed sequence representations are more or less arbitrary. As observed from our experiments (see next Section), this BFS-based model performs equally well (or even slightly better) than those recently proposed models.

\item \textbf{Lexical-Token-Only Model}. This basic attention-based seq2seq model has only one encoder that learns from the lexical information in the source code. We used this model to understand the importance of incorporating syntactic information (in the AST) in deep-learning approaches for code summarization. The parameter setting of the model is similar to that of the Hybrid-DeepCom model.

\end{enumerate}

\subsection{Metrics}
To evaluate the effectiveness of different approaches, we use four widely-used machine translation metrics: two BLEU scores, the METEOR score, and ROUGE-L. The \textbf{BLEU Score} (\citealt{a25}) is a family of metrics to check the quality of machine-translated texts against that of the human-written texts. In this paper, use the Sentence-Level BLEU score (S-BLUE) with smoothing-4 method (\citealt{a26}) and Corpus-level BLEU (C-BLUE) and computed them by a program “multi-bleu.perl”. 
\textbf{METEOR} is a is recall-oriented evaluation method (\citealt{a24}), which evaluates the translation hypotheses by aligning them to reference translations and calculating sentence-level similarity scores (\citealt{a24}).  \textbf{ROUGE-L} (\citealt{ROUGE}) is one of the four measures of the ROUGE family, where L stands for Longest Common Subsequence. It computes the  F-score (defined as the harmonic mean of the recall and precision value from finding the longest common sequence of the texts.)

\subsection{Analysis of Experimental Results and Observations}

In Tables 3 and 4, we summarize the experiment results on the effectiveness of our Pr{\"u}fer-based learning model and the baseline models for code summarization on two public Java datasets: 
a)  Dataset-1 (\citealt{a23}) and b) Dataset-2 (\citealt{codexglue}). 
We note that the performance scores for Dataset-2 is much lower, but this is not a surprise---as we mentioned in the previous section, 
this dataset is believed to be one of the most challenging dataset for deep learning approaches to program understanding and generation.
A detailed discussion can be found on CodeXGlue's project webpage (https://microsoft.github.io/CodeXGLUE/). Also, note that in Table 4 (for Dataset-2), we do not have a row for the model TL-CodeSum because Dataset-2 does not provide the API knowledge required to train the model.  As we can see, the model that uses our Pr{\" u}fer Sequence Encoder and Hu's Source Encoder (second last rows in both Tables 3 and 4) has already had notable improvement over the baseline models. 
The improvement by our complete Pr{\"u}fer-sequence-based model (using our  Pr{\" u}fer Sequence Encoder and our Context Encoder) 
is even more significant, especially on Dataset-2 (last rows in both Tables  3 and 4.)      

In the rest of this section, we discuss our observations from the experiments and analyze the results to understand the power, effectiveness, efficiency, and the robustness (against the code length) of our Pr{\" u}fer-sequence-based  representation. These observations and their analyses are solid evidence, supporting our belief that our  Pr{\" u}fer-sequence-based representation makes it possible (or much easier) for learning models to exploit information in the training examples that are hard, if not impossible, for other recently-proposed learning models to detect. 
%
\subsubsection{Power and Effectiveness of Pr{\"u}fer-Sequence Representation of AST.}  
 
As shown in the second last rows in Tables 3 and 4, our Pr{\" u}fer Sequence Encoder and Hu's Source Encoder (second last rows in both Tables 3 and 4) has already had notable improvement over the baseline models, with the average performance improvement being  $0.9\%$ for Dataset-1 and $3.96\%$ for Dataset-2. We attribute the performance improvement to the properties of the Pr{\"u}fer sequence representation discussed in the previous Section (Abstract Syntax Trees and Pr{\"u}fer Sequences): a concise and lossless encoding that quantifies the ``importance" of syntactic tokens and preserves their structural roles in describing the source code. This is further supported by the relatively poor performance of the recently-proposed general model, Code2Seq, that uses a lossy encoding with the input length cubic to the size of the AST in the worst case. Note that the performance of the Code2Seq Model is worse than the model that does not make use of any structural information of the AST (first rows in both tables.)

We can also see that the performance of the two versions of the Hybrid-DeepCom Model based on respectively the BFS sequence and the SBT sequence are comparable on both datasets, justifying our claim in the introduction that in such traversal-based sequence representation recently proposed, the order of the tokens in a sequence is largely arbitrary and only partially captures the structure of an AST.                  
  
\subsubsection{Importance of Structure-Aware Context Sequence of Lexical Tokens.}   
   
As shown in the last row in both Tables 3 and 4, the use of the Context Encoder boosted the performance of our model dramatically. The average performance improvement over baseline models is increased from $0.9\%$ to $5\%$  for Dataset-1 and $3.96\%$ to $15.11\%$ for Dataset-2. 

Considering that both of the two recently-proposed deep-learning models (Code2Seq and Hybrid-DeepCom,  the second rows and the third last rows in both tables) are also designed to make use of structural information of an AST as well as lexical tokens from the source code, the significant performance gain of our model is best interpreted by the fact that the ``context" sequence defined in our model as the input to the Context Encoder is structure-aware. The difference between our context sequence and the those used in the Hybrid-DeepCom model and the Code2Seq model is that in our context sequence, the frequency of a lexical token from the source code is decided by the degree of its parent node in the AST, whereas Hybrid-DeepCom and Code2Seq treat tokens from the source code equally regardless of their role and significance in the program.          
    
\subsubsection{Efficiency of the Pr{\"u}fer Sequence Encoding.}

The input dimension to a seq-to-seq deep learning model depends on the encoding scheme. The lower the dimension is, the faster it is to complete one epoch of training. There is, of course,  a tradeoff among the effectiveness/ability of a model, its input dimension, and the training time. An ideal encoding is the one that preserves as much as possible the structural information and has a short length.  
 
 \begin{figure}[h!]
\centerline{\includegraphics[width=0.5\textwidth, height=100pt]{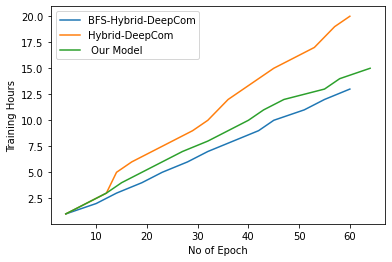}}
\label{fig}
\caption{Time required to train different the models with different AST representation. }
\end{figure}
Our experiments confirmed that our Pr{\" u}fer sequence representation, while encoding the structure of an AST losslessly, is more concise than other representations and indeed requires less training time. Fig. 3 summarizes our observation on the time required to complete different training epochs for three learning models: Our Model,  Hybrid-DeepCom, BFS-Hybrid-DeepCom. Among the models, BFS-Hybrid-DeepCom uses the shortest sequence representation (70.21 on average over the training data), and Hybrid-DeepCom has the longest representation (193.91 on average). The average length of our Pr{\" u}fer sequence representation is 100.81.  It is a surprise to observe that BFS-Hybrid-DeepCom (a model we customized from 
Hybrid-DeepCom using a straightforward and much shorter breadth-first-search-based representation), requires less training time but has a comparable performance with Hybrid-DeepCom that is based on a carefully designed and more sophisticated representation.  While our model requires more time to train than BFS-Hybrid-DeepCom (as expected but not by much), the performance gain of our model is significant. 

\subsubsection{Performance over Source Code of  Different Lengths.} 

\begin{figure}[h!]
  \includegraphics[width=0.5\textwidth,height=3.5cm]{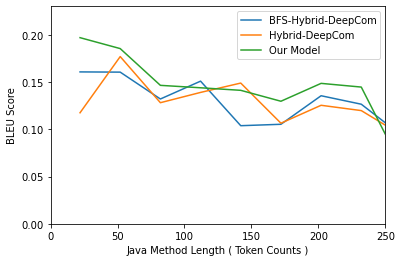}
 \caption{ BLEU score for different method lengths }
\end{figure}
 
We further analyzed the accuracy of the  three models when they are 
trained and tested on source codes of different lengths. We observed (Fig.4) that for all the three models, the performance decreases as the code length increases. We note that our Pr{\"u}fer-sequence-based model performs better than the other two baseline models regardless of the code lengths. For java methods with 150 or more tokens, the Pr{\"u}fer sequence-based model had a clear edge over the other two methods, suggesting that it is more robust against the increase of code length. 

The correlation coefficient between the BLEU score and code length is the smallest for the Pr{\"u}fer-sequence-based model (-0.037), while for BFS and SBT, it is -0.16 and -0.19, respectively. This indicates that the (negative) correlation between the performance of and the code length is much weaker for our model than 
the other two models.


\begingroup
\setlength{\tabcolsep}{1pt} 
\renewcommand{\arraystretch}{1} 
\begin{table}[ht]
\caption{Statistics of the AST Representation Methods for Dataset-1}
  \begin{center}
    \begin{tabular}{l|c|r|r|r} 
      \textbf{Method} & \textbf{Avg.} & \textbf{Mode} &\textbf{Median} &  \textbf{$<$200 Tokens} \\
      \hline
       Prufer Sequence  & 100.81 & 8 & 67.74 & 88.06 \%\\
       SBT  & 193.71 & 24 & 124  & 68.77\% \\
       BFS  & 70.21 & 8 & 44  & 94.16\% \\
 \hline
    \end{tabular}
     \vspace{-1em}
  \end{center}
\end{table}
\endgroup

\section{Conclusion}

In this work, we proposed a concise and effective representation scheme that can be used in sequence-to-sequence models for code representation learning. By encoding structural information of abstract syntax trees of computer programs, our Pr{\"u}fer-sequence-based representation makes it possible (or much easier) to develop sequence-to-sequence learning models to exploit automatically and selectively lexical and syntactic signals that are hard, if not impossible, for other recently-proposed sequence-to-sequence learning models to detect. 


\cite{a9} investigated the possibility of incorporating AST information in their 
transformer-based model and concluded that AST information does not provide any help. 
Our studies in this paper suggest that it depends on how the (hierarchical) AST information is encoded and used in a learning model. It is a very interesting 
future work to study how  our Pr{\" u}fer-sequence-based encoding of ASTs 
can be used in a transformer-based  model (such as the one in \citep{a9}) to 
decrease the model's complexity and improve its effectiveness.      

While the model we developed and the experiments conducted are on the task of code summarization for a particular program language, no assumptions were made about the programming language, its specification, and the format of the abstract syntax trees, making our approach language independent and potentially applicable to other tasks in program comprehension and in any application domains of sequence-to-sequence learning models where sequential and hierarchical signals both exist in the training data.

\bibliographystyle{apalike} 
\bibliography{ref.bib}

\begin{thebibliography}{}

\bibitem[Ahmad et~al., 2020]{a9}
Ahmad, W.~U., Chakraborty, S., Ray, B., and Chang, K.-W. (2020).
\newblock A transformer-based approach for source code summarization.
\newblock In {\em Proceedings of the 58th Annual Meeting of the Association for
  Computational Linguistics (ACL)}.

\bibitem[Allamanis et~al., 2015]{a11}
Allamanis, M., Tarlow, D., Gordon, A., and Wei, Y. (2015).
\newblock Bimodal modelling of source code and natural language.
\newblock In Bach, F. and Blei, D., editors, {\em Proceedings of the 32nd
  International Conference on Machine Learning}, volume~37 of {\em Proceedings
  of Machine Learning Research}, pages 2123--2132, Lille, France. PMLR.

\bibitem[Alon et~al., 2019]{a16}
Alon, U., Brody, S., Levy, O., and Yahav, E. (2019).
\newblock code2seq: Generating sequences from structured representations of
  code.
\newblock In {\em International Conference on Learning Representations}.

\bibitem[Alon et~al., 2018]{a4}
Alon, U., Zilberstein, M., Levy, O., and Yahav, E. (2018).
\newblock A general path-based representation for predicting program
  properties.
\newblock In {\em Proceedings of the 39th ACM SIGPLAN Conference on Programming
  Language Design and Implementation}, PLDI 2018, page 404–419. Association
  for Computing Machinery.

\bibitem[Balog et~al., 2016]{a10}
Balog, M., Gaunt, A.~L., Brockschmidt, M., Nowozin, S., and Tarlow, D. (2016).
\newblock Deepcoder: Learning to write programs.
\newblock {\em CoRR}, abs/1611.01989.

\bibitem[Bielik et~al., 2016]{a13}
Bielik, P., Raychev, V., and Vechev, M. (2016).
\newblock Phog: Probabilistic model for code.
\newblock In Balcan, M.~F. and Weinberger, K.~Q., editors, {\em Proceedings of
  The 33rd International Conference on Machine Learning}, volume~48 of {\em
  Proceedings of Machine Learning Research}, pages 2933--2942. PMLR.

\bibitem[Chen and Cherry, 2014]{a26}
Chen, B. and Cherry, C. (2014).
\newblock A systematic comparison of smoothing techniques for sentence-level
  {BLEU}.
\newblock In {\em Proceedings of the Ninth Workshop on Statistical Machine
  Translation}, pages 362--367, Baltimore, Maryland, USA. Association for
  Computational Linguistics.

\bibitem[Cho and van Merrienboer, 2014]{a21}
Cho, K. and van Merrienboer, B. (2014).
\newblock Learning phrase representations using rnn encoder-decoder for
  statistical machine translation.
\newblock In {\em Proceedings of the 2014 Conference on Empirical Methods in
  Natural Language Processing (EMNLP)}, pages 1724--1734.

\bibitem[Denkowski and Lavie, 2014]{a24}
Denkowski, M. and Lavie, A. (2014).
\newblock Meteor universal: Language specific translation evaluation for any
  target language.
\newblock In {\em Proceedings of the Ninth Workshop on Statistical Machine
  Translation}, pages 376--380, Baltimore, Maryland, USA. Association for
  Computational Linguistics.

\bibitem[Haije et~al., 2016]{a1}
Haije, T., Intelligentie, B. O.~K., Gavves, E., and Heuer, H. (2016).
\newblock Automatic comment generation using a neural translation model.
\newblock {\em Inf. Softw. Technol}, 55(3):258--268.

\bibitem[Haque et~al., 2020]{a8}
Haque, S., LeClair, A., Wu, L., and McMillan, C. (2020).
\newblock Improved automatic summarization of subroutines via attention to file
  context.
\newblock {\em Proceedings of the 17th International Conference on Mining
  Software Repositories}.

\bibitem[Hashemi and Tari, 2018]{pru-r2}
Hashemi, Z. and Tari, F.~G. (2018).
\newblock A prufer-based genetic algorithm for allocation of the vehicles in a
  discounted transportation cost system.
\newblock {\em International Journal of Systems Science: Operations \&
  Logistics}, 5(1):1--15.

\bibitem[Hu et~al., 2018a]{a5}
Hu, X., Li, G., Xia, X., Lo, D., and Jin, Z. (2018a).
\newblock Deep code comment generation.
\newblock In {\em Proceedings of the 26th Conference on Program Comprehension},
  ICPC '18, page 200–210. Association for Computing Machinery.

\bibitem[Hu et~al., 2020]{a15}
Hu, X., Li, G., Xia, X., Lo, D., and Jin, Z. (2020).
\newblock Deep code comment generation with hybrid lexical and syntactical
  information.
\newblock {\em Empirical Software Engineering}, 25(3):2179--2217.

\bibitem[Hu et~al., 2018b]{a23}
Hu, X., Li, G., Xia, X., Lo, D., Lu, S., and Jin, Z. (2018b).
\newblock Summarizing source code with transferred api knowledge.
\newblock In {\em Proceedings of the 27th International Joint Conference on
  Artificial Intelligence}, IJCAI'18, page 2269–2275. AAAI Press.

\bibitem[Iyer et~al., 2016]{a6}
Iyer, S., Konstas, I., Cheung, A., and Zettlemoyer, L. (2016).
\newblock Summarizing source code using a neural attention model.
\newblock In {\em Proceedings of the 54th Annual Meeting of the Association for
  Computational Linguistics (Volume 1: Long Papers)}, pages 2073--2083.
  Association for Computational Linguistics.

\bibitem[LaToza et~al., 2006]{b2}
LaToza, T.~D., Venolia, G., and DeLine, R. (2006).
\newblock Maintaining mental models: a study of developer work habits.
\newblock In {\em Proceedings of the 28th international conference on Software
  engineering}, pages 492--501.

\bibitem[Lin, 2004]{ROUGE}
Lin, C.-Y. (2004).
\newblock {ROUGE}: A package for automatic evaluation of summaries.
\newblock In {\em Text Summarization Branches Out}, pages 74--81, Barcelona,
  Spain. Association for Computational Linguistics.

\bibitem[Lu et~al., 2021]{codexglue}
Lu, S., Guo, D., Ren, S., Huang, J., Svyatkovskiy, A., Blanco, A., Clement,
  C.~B., Drain, D., Jiang, D., Tang, D., Li, G., Zhou, L., Shou, L., Zhou, L.,
  Tufano, M., Gong, M., Zhou, M., Duan, N., Sundaresan, N., Deng, S.~K., Fu,
  S., and Liu, S. (2021).
\newblock Codexglue: {A} machine learning benchmark dataset for code
  understanding and generation.
\newblock {\em CoRR}, abs/2102.04664.

\bibitem[Molla-Alizadeh-Zavardehi et~al., 2011]{pru-r1}
Molla-Alizadeh-Zavardehi, S., Hajiaghaei-Keshteli, M., and Tavakkoli-Moghaddam,
  R. (2011).
\newblock Solving a capacitated fixed-charge transportation problem by
  artificial immune and genetic algorithms with a pr\"{u}fer number
  representation.
\newblock 38(8):10462–10474.

\bibitem[Nayeem and Pal, 2013]{pru-r3}
Nayeem, S. M.~A. and Pal, M. (2013).
\newblock Diameter constrained fuzzy minimum spanning tree problem.
\newblock {\em International Journal of Computational Intelligence Systems},
  6(6):1040--1051.

\bibitem[Papineni et~al., 2002]{a25}
Papineni, K., Roukos, S., Ward, T., and Zhu, W.-J. (2002).
\newblock {B}leu: a method for automatic evaluation of machine translation.
\newblock In {\em Proceedings of the 40th Annual Meeting of the Association for
  Computational Linguistics}, pages 311--318, Philadelphia, Pennsylvania, USA.
  Association for Computational Linguistics.

\bibitem[Pradhan and Bhattacharya, 2019]{pru-r8}
Pradhan, M. and Bhattacharya, B.~B. (2019).
\newblock A prufer-sequence based representation of large graphs for structural
  encoding of logic networks.
\newblock In {\em Proceedings of the ACM India Joint International Conference
  on Data Science and Management of Data}, CoDS-COMAD '19, page 293–296, New
  York, NY, USA. Association for Computing Machinery.

\bibitem[Pr{\"u}fer, 1918]{a18}
Pr{\"u}fer, H. (1918).
\newblock Neuer beweis eines satzes {\"u}ber permutationen (a new prof. of a
  theorem on permutations).
\newblock {\em Archiv der mathematik und Physik}, 3:27.

\bibitem[Raychev et~al., 2016a]{a14}
Raychev, V., Bielik, P., and Vechev, M. (2016a).
\newblock Probabilistic model for code with decision trees.
\newblock In {\em Proceedings of the 2016 ACM SIGPLAN International Conference
  on Object-Oriented Programming, Systems, Languages, and Applications}, OOPSLA
  2016, page 731–747. Association for Computing Machinery.

\bibitem[Raychev et~al., 2016b]{a12}
Raychev, V., Bielik, P., Vechev, M., and Krause, A. (2016b).
\newblock Learning programs from noisy data.
\newblock In {\em Proceedings of the 43rd Annual ACM SIGPLAN-SIGACT Symposium
  on Principles of Programming Languages}, POPL '16, page 761–774, New York,
  NY, USA. Association for Computing Machinery.

\bibitem[Sridhara et~al., 2010]{b3}
Sridhara, G., Hill, E., Muppaneni, D., Pollock, L., and Vijay-Shanker, K.
  (2010).
\newblock Towards automatically generating summary comments for java methods.
\newblock In {\em Proceedings of the IEEE/ACM international conference on
  Automated software engineering}, pages 43--52.

\bibitem[Sutskever et~al., 2014]{a19}
Sutskever, I., Vinyals, O., and Le, Q.~V. (2014).
\newblock Sequence to sequence learning with neural networks.
\newblock In {\em Proceedings of the 27th International Conference on Neural
  Information Processing Systems - Volume 2}, NIPS'14, page 3104–3112. MIT
  Press.

\bibitem[Wan et~al., 2018]{a7}
Wan, Y., Zhao, Z., Yang, M., Xu, G., Ying, H., Wu, J., and Yu, P.~S. (2018).
\newblock Improving automatic source code summarization via deep reinforcement
  learning.
\newblock {\em CoRR}, abs/1811.07234.

\bibitem[West, 2000]{z1}
West, D.~B. (2000).
\newblock {\em Introduction to Graph Theory}.
\newblock Prentice Hall.

\bibitem[Xu et~al., 2008]{pru-r5}
Xu, J., Liu, Q., and Wang, R. (2008).
\newblock A class of multi-objective supply chain networks optimal model under
  random fuzzy environment and its application to the industry of chinese
  liquor.
\newblock {\em Inf. Sci.}, 178(8):2022–2043.

\bibitem[Yang and Wang, 2020]{pru-r7}
Yang, L. and Wang, Y. (2020).
\newblock Prufer coding: A vectorization method for undirected labeled graph.
\newblock {\em IEEE Access}, 8:175360--175369.

\bibitem[Zhang et~al., 2020]{a3}
Zhang, J., Wang, X., Zhang, H., Sun, H., and Liu, X. (2020).
\newblock Retrieval-based neural source code summarization.
\newblock In {\em Proceedings of the ACM/IEEE 42nd International Conference on
  Software Engineering}, ICSE '20, page 1385–1397. Association for Computing
  Machinery.

\end{thebibliography}

\end{document}